\documentclass[runningheads]{llncs}
\usepackage{graphicx}
\usepackage{booktabs, multirow}
\usepackage[misc]{ifsym}

\newcommand{\ignore}[1]{}

\begin{document}
\title{Recurrent Connections Aid Occluded Object Recognition by Discounting Occluders\thanks{This work was supported by the European Union’s Horizon 2020 research and innovation programme under grant agreement N\textsuperscript{\underline{o}} 713010  (GOAL-Robots, Goal-based Open-ended Autonomous Learning Robots).}}
\titlerunning{Recurrent Connections Aid Occluded Object Recognition}
%
\author{Markus Roland Ernst \inst{1,2}\ignore{ \orcidID{0000-0002-2800-6346} }\textsuperscript{(\Letter)} \and
Jochen Triesch \inst{1,2}\ignore{ \orcidID{0000-0001-8166-2441} } \and
Thomas Burwick \inst{1,2}\ignore{ \orcidID{0000-0002-2019-4119} } }
\authorrunning{M. R. Ernst et al.}

\institute{Frankfurt Institute for Advanced Studies, \\ Ruth-Moufang-Stra\ss e 1, 60438 Frankfurt am Main, Germany \and Goethe-Universit\"at Frankfurt, \\ Max-von-Laue-Stra\ss e 1, 60438 Frankfurt am Main, Germany\\
\email{\{mernst,triesch,burwick\}@fias.uni-frankfurt.de}}
\maketitle 
\begin{abstract}
Recurrent connections in the visual cortex are thought to aid object recognition when part of the stimulus is occluded. Here we investigate if and how recurrent connections in artificial neural networks similarly aid object recognition. We systematically test and compare architectures comprised of bottom-up (B), lateral (L) and top-down (T) connections. Performance is evaluated on a novel stereoscopic occluded object recognition dataset. The task consists of recognizing one target digit occluded by multiple occluder digits in a pseudo-3D environment. We find that recurrent models perform significantly better than their feedforward counterparts, which were matched in parametric complexity. Furthermore, we analyze how the network's representation of the stimuli evolves over time due to recurrent connections. We show that the recurrent connections tend to move the network's representation of an occluded digit towards its un-occluded version. Our results suggest that both the brain and artificial neural networks can exploit recurrent connectivity to aid occluded object recognition.

\keywords{Object recognition \and Occlusion \and Recurrent neural networks.}
\end{abstract}
\section{Introduction}
Given the rapidness of invariant object recognition in primates \cite{thorpe1996speed,isik2013dynamics}, the process is assumed to be mostly feedforward \cite{dicarlo2012does}. This assumption has been corroborated by the recent success of feedforward neural networks in computer vision \cite{krizhevsky2012imagenet,lecun2015deep} and led to modelling of the primate visual system using such networks \cite{riesenhuber1999hierarchical,serre2007feedforward}. 
However, both anatomical and functional evidence suggest that recurrent connections do indeed influence object recognition. Densities of feedforward and recurrent connections in the ventral visual pathway are comparable in magnitude \cite{felleman1991distributed,sporns2004small} and electrophysiological experiments have demonstrated that the processing of object information unfolds over time, beyond what would normally be attributed to a feedforward process \cite{cichy2014resolving,brincat2006dynamic}.
In particular, recognition of degraded or occluded objects produces delayed behavioral and neural responses \cite{johnson2005recognition,tang2014spatiotemporal} believed to be caused by competitive processing due to lateral recurrent connections \cite{adesnik2012lateral}.

Other evidence suggests that recurrent top-down connections can fill in missing information in partially occluded images \cite{oreilly2013recurrent}. Additionally, recurrent convolutional neural networks have been shown to improve classification performance on occluded stimuli \cite{spoerer2017recurrent,liang2015recurrent}. 
However, the stimuli used in previous research did hardly resemble an authentic natural environment. The world humans live in and interact with is inherently 3D and occlusion consists of more than just masking one stimulus with another. Rather, it is highly dependent on viewing angle, and primates perceive it stereoscopically with two eyes.
Unlike previous simulations and experimental work, where part of the input image was deleted or masked in two dimensions \cite{oreilly2013recurrent,wyatte2012limits,tang2014spatiotemporal,spoerer2017recurrent}, we set out to test the effects of occlusion in a more natural environment. 
Thus, we extended the generative model for occluded stimuli presented in \cite{spoerer2017recurrent} to account for 3D perspective and stereo vision.

We test and compare a range of different recurrent convolutional neural network architectures, assuming the naming scheme of \cite{spoerer2017recurrent,liang2015recurrent}. Bottom-up (B) and top-down (T) connections correspond to processing information from lower and higher regions in the ventral visual hierarchy, and lateral (L) connections process information within a region.

To investigate whether recurrent networks outperform feedforward models in a more naturalistic setting, the different architectures were tasked with classifying objects under varying levels of occlusion. The accuracy or error rate reflects the degree to which the networks learn to recognize the target, and how well they cope with occlusion. Additionally we train and test all networks on stereoscopic image data to quantify the benefit of binocular vision.
Finally we explore how recurrent connections shape the probability distribution over possible outcomes and we analyze the internal representation of the occluded stimuli. We conduct a geometrical analysis of activations in the final hidden layer and visualize the evolution of the internal representation in time using t-distributed stochastic neighbor embedding (t-SNE) \cite{maaten2008visualizing}. Our results demonstrate significant performance advantages of recurrent networks and reveal how recurrence helps to discount the effect of occluders.

\section{Methods}
\subsection{Stereo-Digits Dataset}
We investigate the effects of occlusion using a novel stereoscopic image data set. Inspired by the generative image model for occlusion stimuli in \cite{spoerer2017recurrent}, we focus on digit recognition. Our stereo-digits dataset is meant to bridge the gap between the somewhat artificial task of recognizing computer rendered digits and the natural task of recognizing partially occluded objects. Contrary to past studies, occlusion is generated by overlapping the target stimulus with other digit instances in a pseudo-3D environment.

\begin{figure}[hbtp]
\centering
\includegraphics[width=0.49\textwidth]{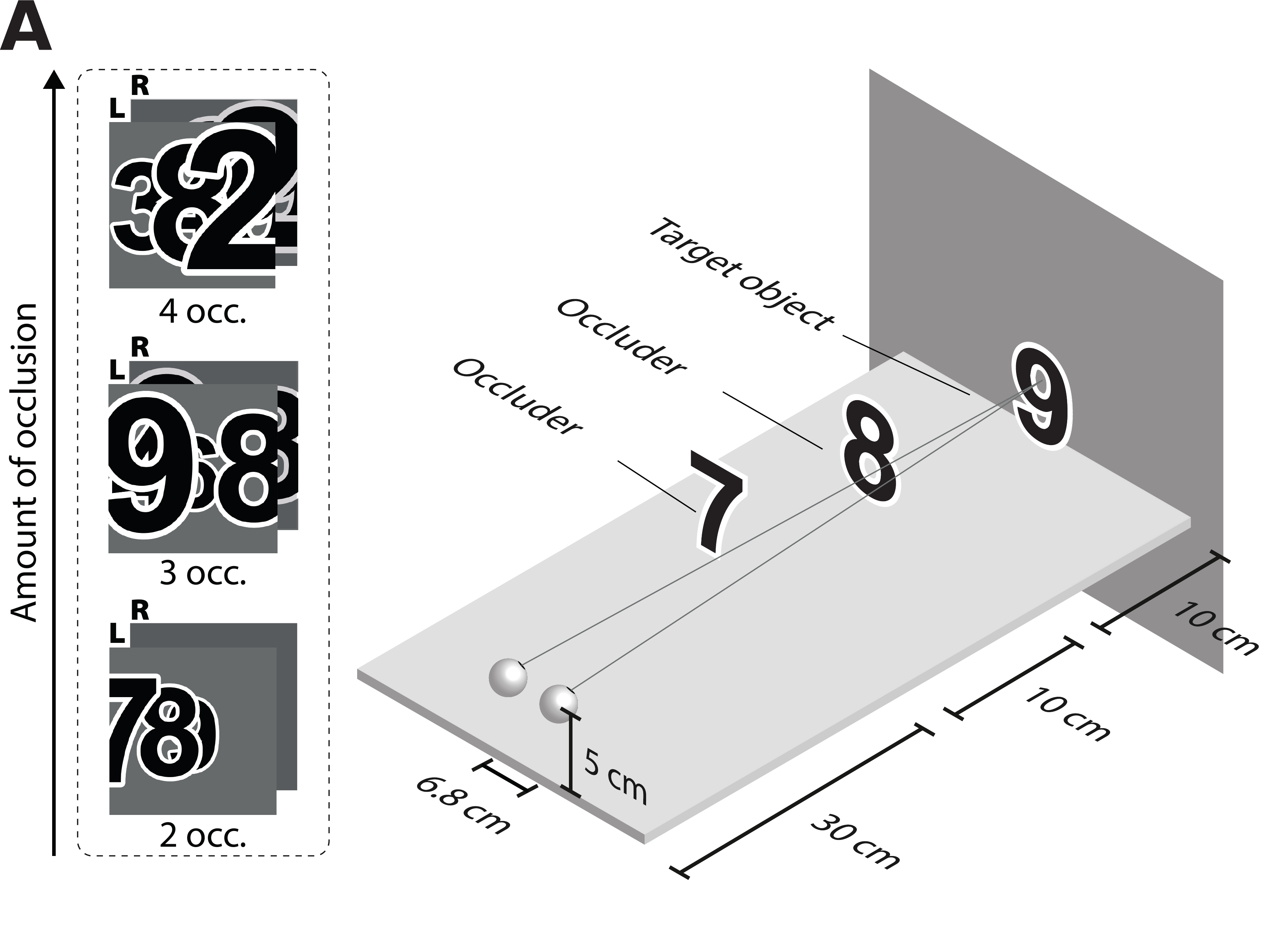}
\includegraphics[width=0.49\textwidth]{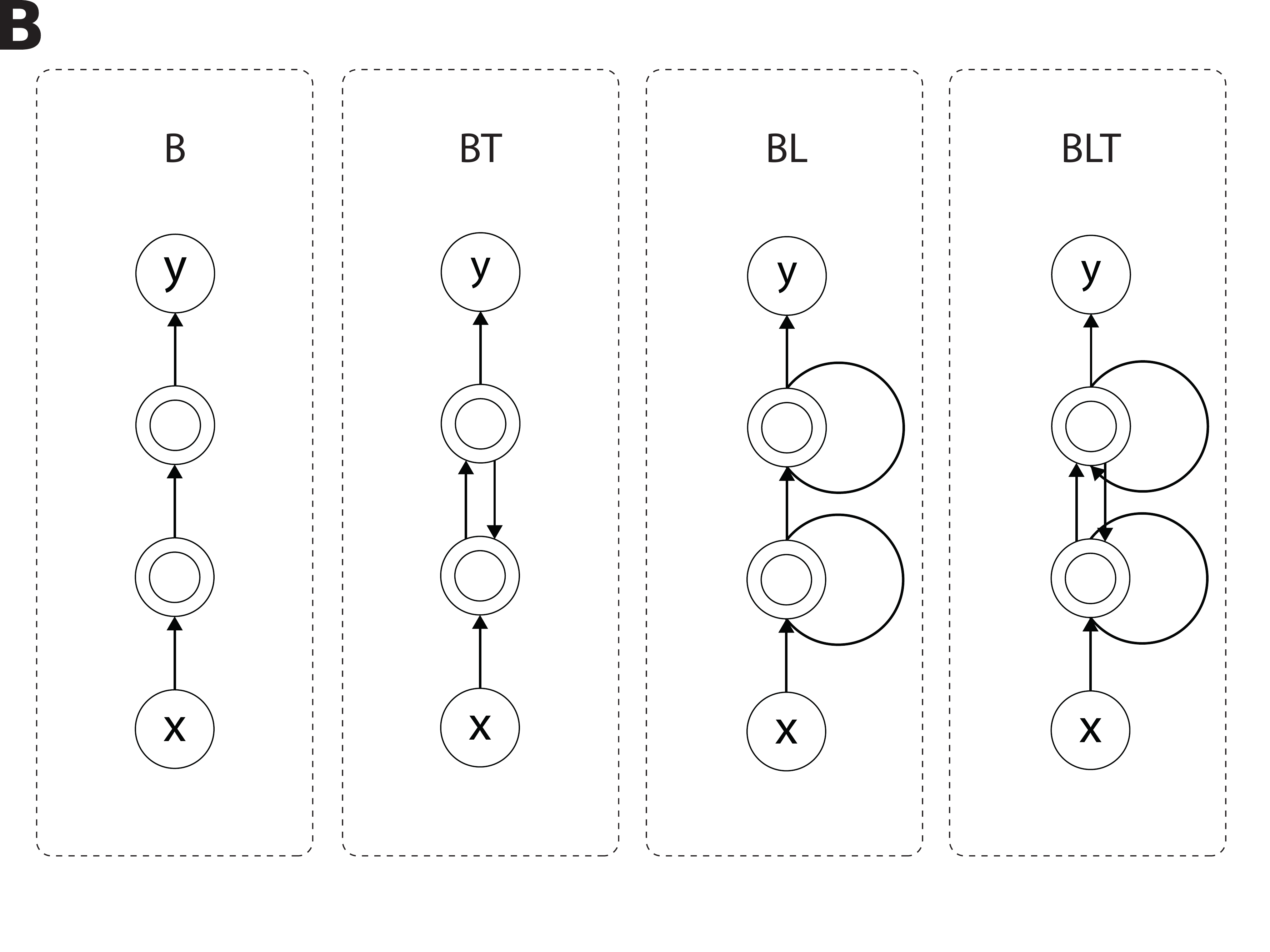}
\caption{The used stimuli and network models. (A) The centered target object is occluded by 2--4 digits arranged in a 3D-fashion. (B) A sketch of the four network architectures named after their connection properties. B stands for bottom-up, L for lateral and T for top-down connections.}
\label{fig:network_overview}
\end{figure}

All images of the stereo-digits dataset contain digits of the same font and color. Occlusion is generated by overlaying digits on top of each other as shown in Fig.~\ref{fig:network_overview} A.
The target object, i.e. the hindmost digit, is centered in the middle of the square canvas. Additional digits are then sequentially placed on top of the target object. These occluding objects remain fixed along the y-axis as if standing on a surface 5 cm below the viewer. The x-coordinate is drawn from a uniform distribution. The font size of the digits was scaled to give the impression of objects with 20 cm height placed at different depths. We assumed a distance of 50 cm from the target object to the viewer, and 10 cm less for every added object. The level of occlusion can be controlled by varying the number of occluder objects, which increases the difficulty of the task. Images for the left and right eye were taken given an interocular distance of 6.8 cm. 
Each dataset (2, 3, 4 occluders) consists of 100,000 randomly generated images for training and 10,000 images for testing. The images were rendered at a resolution of $512 \times 512$ and then downsampled to $32 \times 32$.

\subsection{Network Models}
Four basic network models were compared as shown in Fig.~\ref{fig:network_overview} B: Bottom-up connection only (\emph{B}), bottom-up and top-down connections (\emph{BT}), bottom-up and lateral connections (\emph{BL}), and bottom-up, lateral, and top-down connections (\emph{BLT}). As lateral and top-down connections introduce cycles into the computational graph, these models represent recurrent neural networks and allow for information to be retained within a layer or to flow back into earlier layers.

Each model consists of an input layer, two hidden recurrent layers and an output layer. Both bottom-up and lateral connections are implemented as convolutional layers \cite{lecun1998gradient} with a stride of $1 \times 1$. After convolution the activations go through a $2 \times 2$ maxpooling layer with a stride of $2 \times 2$.
The top-down connections are implemented as a transposed convolution \cite{zeiler2010deconvolutional} with output stride $2 \times 2$ to match the input size of the convolutional layer that came before it. Each of the recurrent network models is unrolled and trained for four time steps by backpropagation \cite{rumelhart1986learning}. When measuring accuracy, the output at the final unrolled time step available for the particular architecture is used.
To compensate for the fact that recurrent network models have more learnable parameters than their non-recurrent counterparts, we introduce two additional feedforward models \emph{B-F} and \emph{B-K}. \emph{B-F} is a feedforward model where the number of convolutional filters or kernels in the hidden layers is increased from 32 to 64. \emph{B-K} has an increased  convolutional kernel size of $5 \times 5$ compared to $3 \times 3$ of the standard \emph{B} model. As a larger kernel effectively increases the number of connections that each unit has, \emph{B-K} is a more appropriate model for control. \emph{B-F} on the other hand alters the representational power of the model by adding more feature maps. The number of learnable parameters for each of the models can be found in Table \ref{tab:networkparameters_fm1o10}.

\begin{table}[hbt]
\centering
\caption{Number of learnable parameters for all models and input channels.}
\begin{tabular}{ccccccc}
\toprule
 & \emph{B} & \emph{B-F}  & \emph{B-K} & \emph{BT} & \emph{BL} &  \emph{BLT} \\
\midrule
Kernel size & $3 \times 3 $ &  $3 \times 3 $      &  $5 \times 5 $    &   $3 \times 3 $ &  $3 \times 3 $ &  $3 \times 3 $  \\
Hidden layer units  & $32$ & $64$      & $32$    &  $32$ & $32$ & $32$  \\
\midrule
&\multicolumn{6}{c}{} \\
Image channels  &\multicolumn{6}{c}{Number of learnable parameters} \\
\cmidrule{0-0} \cmidrule{2-7}
1 & $9{,}898$ & $38{,}218$      & $26{,}794$    & $19{,}146$ &  $28{,}394$  & $37{,}642$  \\
2 & $10{,}186$ & $38{,}794$      & $27{,}594$    & $19{,}434$ &  $28{,}682$  & $37{,}930$  \\
\bottomrule
\end{tabular}

\label{tab:networkparameters_fm1o10}
\end{table}

\subsubsection{Layers}
After the stimulus enters the network, its activations pass two hidden convolutional layers. The inputs to these layer are denoted by $\mathbf{h}_{i,j}^{(t,l)}$. This formulation represents the vectorized input of a patch centered on location $(i,j)$ in layer $l$ computed at time step $t$ across all feature maps indexed by $k$. Assuming this notation the input stimulus presented to the network becomes $\mathbf{h}_{i,j}^{(t,0)}$.
The activation $z$ of a hidden recurrent layer can then be written as
\begin{equation}
	z^{(t,l)}_{i,j,k} =\left( \mathbf{w}_{k}^{(l)B} \right)^\top \mathbf{h}^{(t,l-1)}_{i,j} + \left( \mathbf{w}_{k}^{(l)L} \right)^\top \mathbf{h}^{(t-1,l)}_{i,j} + \left( \mathbf{w}_{k}^{(l)T} \right)^\top \mathbf{h}^{(t-1,l+1)}_{i,j},
\end{equation}
where $\mathbf{w}_{k}^{(l)\cdot}$ is the vectorized form of the convolutional kernel at feature map $k$ in layer $l$ for bottom-up (B), lateral (L), and top-down (T) connections, respectively. These kernels become only active for architectures using the particular connection and are otherwise zero. Note that the lateral and top-down connections depend on values one time step earlier, so the inputs are defined to be a vector of zeroes for $t=0$ where there would be no previous time step. Top-down connections are only present between the two hidden layers (Fig.~\ref{fig:network_overview} B).

Following the flow of information, the $z^{(t,l)}_{i,j,k}$ of the hidden layers are then batch-normalized \cite{ioffe2015batch}. This technique normalizes an activation $z$ using the mean $\mu_\mathcal{B}$ and standard deviation $\sigma_\mathcal{B}$ over a mini-batch of activations $\mathcal{B}$ and adds multiplicative and additive noise. 
\begin{equation}
	\mathrm{BN}_{\gamma, \beta}(z^{(t,l)}_{i,j,k})= \mathbf{\gamma}^{(l)}_k \cdot  \frac{z^{(t,l)}_{i,j,k} - \mathbf{\mu_\mathcal{B}}}{\mathbf{\sigma_\mathcal{B}}} + \mathbf{\beta}^{(l)}_k,
\end{equation}
where $\gamma$ and $\beta$ are additional learnable parameters. 

The output then is passed to rectified linear units (ReLU, $\sigma_z$) 
\begin{equation}
    \sigma_{z} \left(z^{(t,l)}_{i,j,k} \right) = \max \left( 0, z^{(t,l)}_{i,j,k} \right)
\end{equation}
and goes through local response normalization (LRN, $\omega$)
\begin{equation}
	\omega(a^{(t,l)}_{i,j,k}) = a^{(t,l)}_{i,j,k} \left( c + \alpha \sum_{k' = \max(0, k-\frac{n}{2})}^{\min(n-1, k+\frac{n}{2})} \left(a^{(t,l)}_{i,j,k'}\right)^2 \right)^{-\beta},
\end{equation}
with $n=5$, $c=1$, $\alpha = 10^{-4}$ and $\beta = 0.5$. Similar in justification to maxpooling, LRN implements a form of lateral inhibition by inducing competition for large activities amongst outputs computed using different kernels \cite{krizhevsky2012imagenet}.
Finally the output $h^{(t,l)}_{i,j,k}$ for each hidden layer can be written as:
\begin{equation}
	h^{(t,l)}_{i,j,k} = \omega \left( \sigma_z \left( \mathrm{BN}_{\mathbf{\gamma},\mathbf{\beta}} \left( z^{(t,l)}_{i,j,k}\right) \right) \right).
\end{equation}

After the second hidden layer the information flows through a fully-connected segment with ten output units and softmax activation, defined as:
\begin{equation}
	\mathrm{softmax}(\mathbf{a})_i = \frac{\exp(a_i)}{\sum_j \exp(a_j)}.
\end{equation}
The resulting network output can be interpreted as the probability distribution over the ten classes.

\subsubsection{Learning}
The labels to be predicted by the network are encoded as one-hot vectors. 
To make the networks' output $\hat{\mathbf{y}}^{(\tau)}$ match the target $\mathbf{y}$ we use the cross-entropy cost-function summed across all $\tau$ time steps and all $N$ output units:
\begin{equation}
	J(\hat{\mathbf{y}}^{(\tau)}, \mathbf{y}) = - \sum_{t=0}^{\tau} \sum_{i=0}^{N} y_i \cdot \log \hat{y}^{(t)}_i + (1-y_i) \cdot \log(1- \hat{y}^{(t)}_i).
\end{equation}
The Adam algorithm \cite{kingma2014adam} with an initial learning rate of $\eta = 0.003$  was used to perform gradient descent. Unless stated otherwise training occurred for 25 epochs with mini-batches of size 400.

\subsection{Model Performance Metrics and Evaluation Techniques}
The different models were evaluated in terms of classification accuracy averaged across the test set. Test performances were compared with each other using pair-wise McNemar's tests \cite{mcnemar1947note} as suggested in \cite{dietterich1998approximate}. 
This technique does not require repeated training and therefore poses a computationally efficient method to evaluate a variety of different models. 
As multiple comparisons increase the chance of false positives a Bonferroni-type correction procedure was employed to control the false discovery rate (FDR) at 0.05 \cite{benjamini1995controlling}.

\section{Results}
\subsection{Performance Evaluation}

\begin{figure}[hbtp]
\centering
\includegraphics[width=0.49\textwidth]{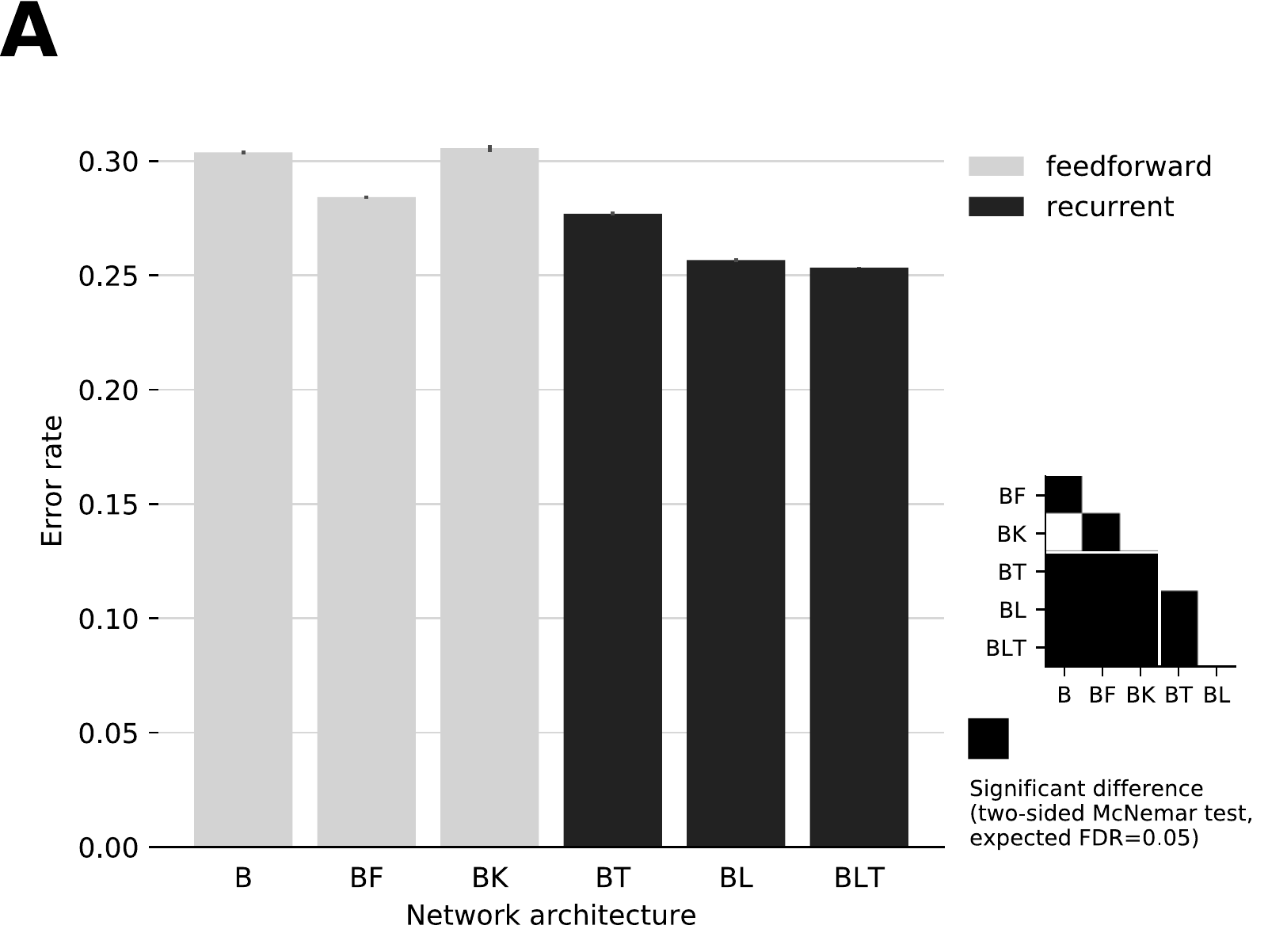}
\includegraphics[width=0.49\textwidth]{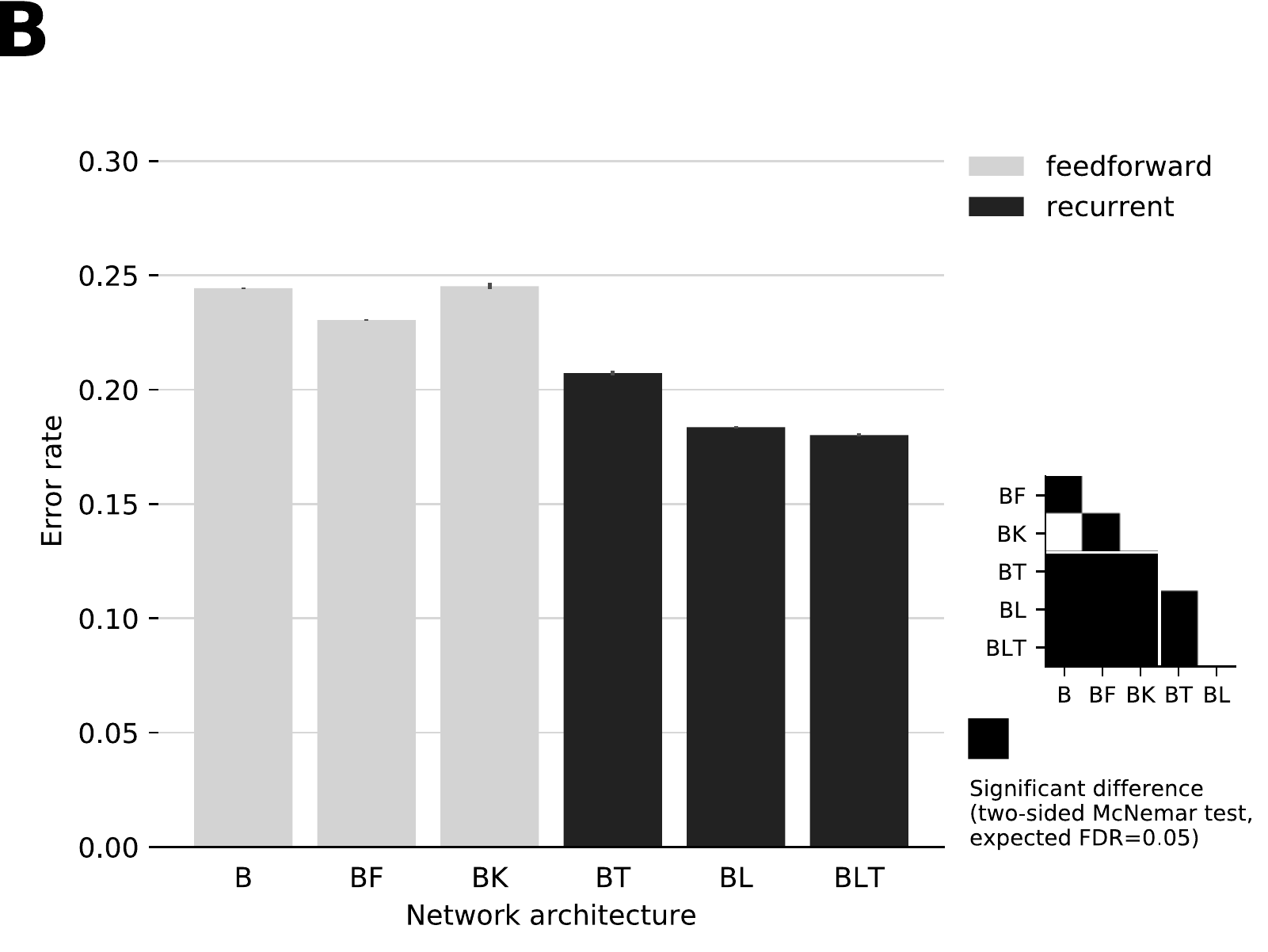}
\caption{Performance comparison of different network architectures. Error bars indicate the standard error based on five repetitions of the training and testing procedure. Matrices depict results of pairwise McNemar tests,  black squares indicating significant differences at $p < 0.05$. (A) Monocular input. (B) Stereoscopic input.}
\label{fig:performance_evaluation}
\end{figure}

Networks were trained on datasets combining all three occlusion levels to evaluate the benefit of feedback connections. Training lasted twenty-five epochs.
Fig.~\ref{fig:performance_evaluation} depicts the classification error $E_{cl}= (1 - \textrm{accuracy})$ for the models trained with monocular (A) and stereoscopic (B) input.

Our results reveal that recurrent architectures perform consistently better than feedforward networks of approximately equal complexity. Notably, \emph{B-K} performs significantly worse than \emph{B-F} questioning the benefits of the increased kernel size, $\chi^2(1,N = 30{,}000) = 46.29, p < .01$.
Significant differences (FDR = 0.05) can be attested for all combinations except (\emph{B}, \emph{B-K}), $\chi^2(1,N = 30{,}000) = 1.69, p = .13$ and (\emph{BL}, \emph{BLT}), $\chi^2(1,N = 30{,}000) = 0.94, p = .26$. The lower left $3 \times 3$ square, highlighted by a white line, indicates that all pair-wise tests between feedforward and recurrent models show a significant advantage of the recurrent architectures. The relative differences in error-rate between feedforward and recurrent models are increased for the stereoscopic case.

When trained separately on the three datasets, we observe almost the same patterns while the error-rates grow with the number of occluders as expected (see Table~\ref{tab:performance_comparison_sdd}). The \emph{BLT} model produces the lowest error-rates for each data set.

\begin{table}[hbt]
\centering
\caption{Error-rates for all model architectures, standard error based on five independent training runs. 2, 3, 4 occ. runs were trained for 100 ep., batchsize 100. Best performance per dataset is highlighted in bold.}
\begin{tabular}{cccccccc}
\toprule
Channels & Occ. & \emph{B} & \emph{B-F}  & \emph{B-K} & \emph{BT} & \emph{BL} &  \emph{BLT} \\
\midrule

\multirow{3}{*}{1 (mono)}
&2& $.134 \pm .004$ & $.123 \pm .003$      & $.143 \pm .002$    &  $.109 \pm .003$ & $.103 \pm .003$ & $\mathbf{.095 \pm .002}$  \\
&3& $.330 \pm .005$ & $.337 \pm .004$      & $.359 \pm .004$    &  $.293 \pm .003$ & $.282 \pm .005$ & $\mathbf{.280 \pm .003}$   \\
&4& $.512 \pm .005$ & $.519 \pm .005$      & $.546 \pm .005$    &  $.477 \pm .003$ & $.463 \pm .005$ & $\mathbf{.455 \pm .006}$   \\
&all& $.304 \pm .001$ & $.284 \pm .001$      & $.306 \pm .002$    &  $.277 \pm .001$ & $.257 \pm .001$ & $\mathbf{.253 \pm .000}$   \\
\cmidrule{1-8}
\multirow{3}{*}{2 (stereo)}
&2& $.095 \pm .003$ & $.078 \pm .003$      & $.094 \pm .003$    &  $.069 \pm .003$ & $.059 \pm .002$ & $\mathbf{.056 \pm .002}$   \\
&3& $.267 \pm .006$ & $.279 \pm .003$      & $.287 \pm .004$    &  $.217 \pm .004$ & $.199 \pm .003$ & $\mathbf{.189 \pm .003}$   \\
&4& $.455 \pm .006$ & $.472 \pm .004$      & $.482 \pm .005$    &  $.395 \pm .003$ & $.373 \pm .003$ & $\mathbf{.361 \pm .003}$  \\
&all& $.244 \pm .000$ & $.230 \pm .000$      & $.245 \pm .001$    &  $.207 \pm .001$ & $.184 \pm .000$ & $\mathbf{.180 \pm .001}$   \\

\bottomrule
\end{tabular}

\label{tab:performance_comparison_sdd}
\end{table}

\subsection{Evolution in Time and Hidden Representation}

\begin{figure}
\includegraphics[width=\textwidth]{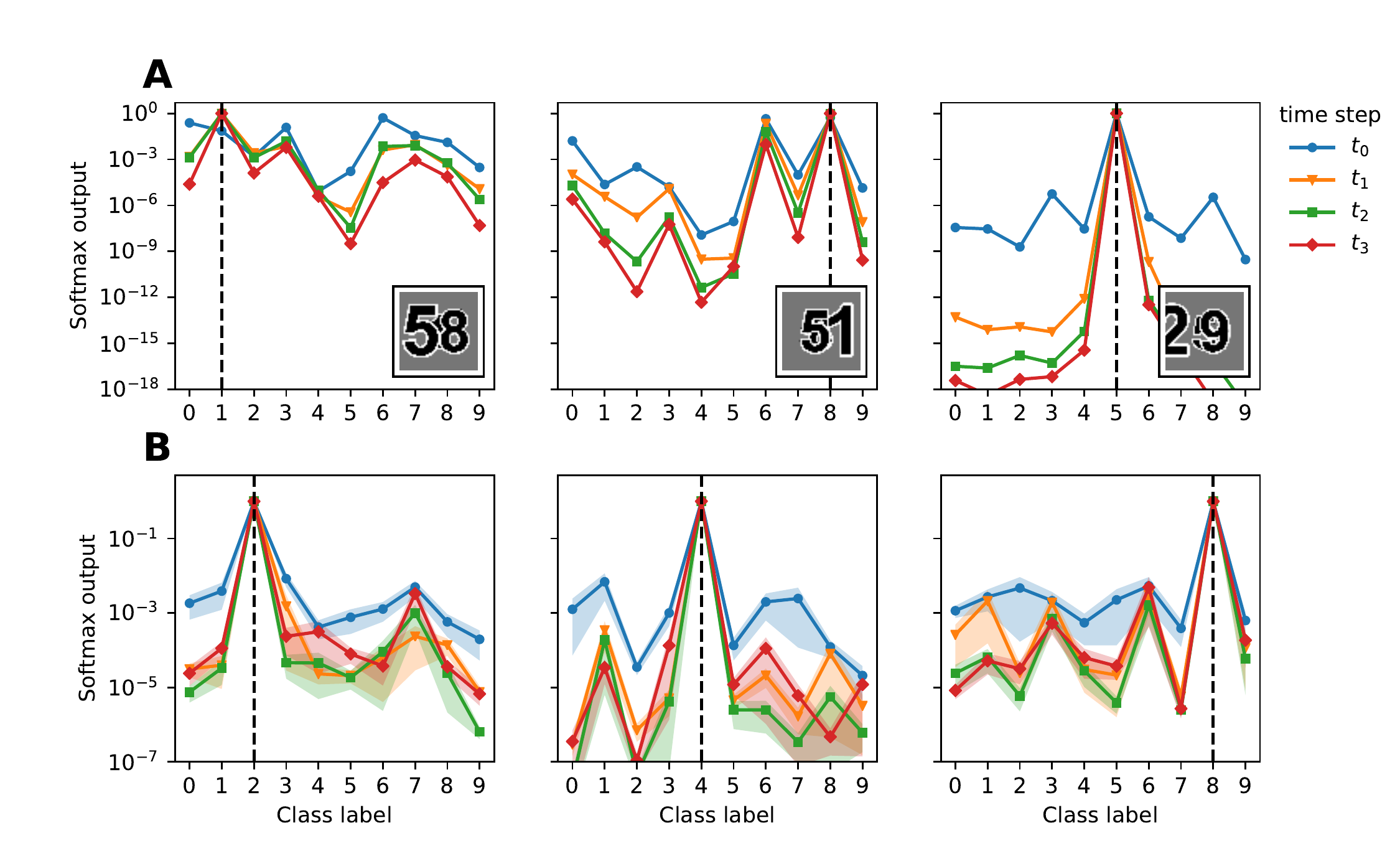}
\caption{Softmax output of \emph{BLT}. (A) Specific stimuli (1, 8, 5) illustrating the effect of recurrent feedback. (B) Mean softmax output over all test stimuli of specific classes (2, 4, 8). Shaded areas correspond to standard error.} 
\label{fig:softmax}
\end{figure}

The softmax output of the \emph{BLT} architecture illustrates how recurrent feedback can revise the network's belief over time. In fact, we observe that wrong initial guesses are being corrected and correct guesses are reinforced. Specific examples are shown in Fig.~\ref{fig:softmax} A: While the network estimates the target digit to be 6 at $t_0$, the final output is the correct answer 1 (left panel). 
The mean softmax activations for specific classes, Fig.~\ref{fig:softmax} B, indicate that the probabilities assigned to incorrect classes decrease over time. Additionally this visualization reveals systematic visual similarities that the network has discovered between digits 2 and 7 and digits 4 and 1.

\begin{figure}
\includegraphics[width=\textwidth]{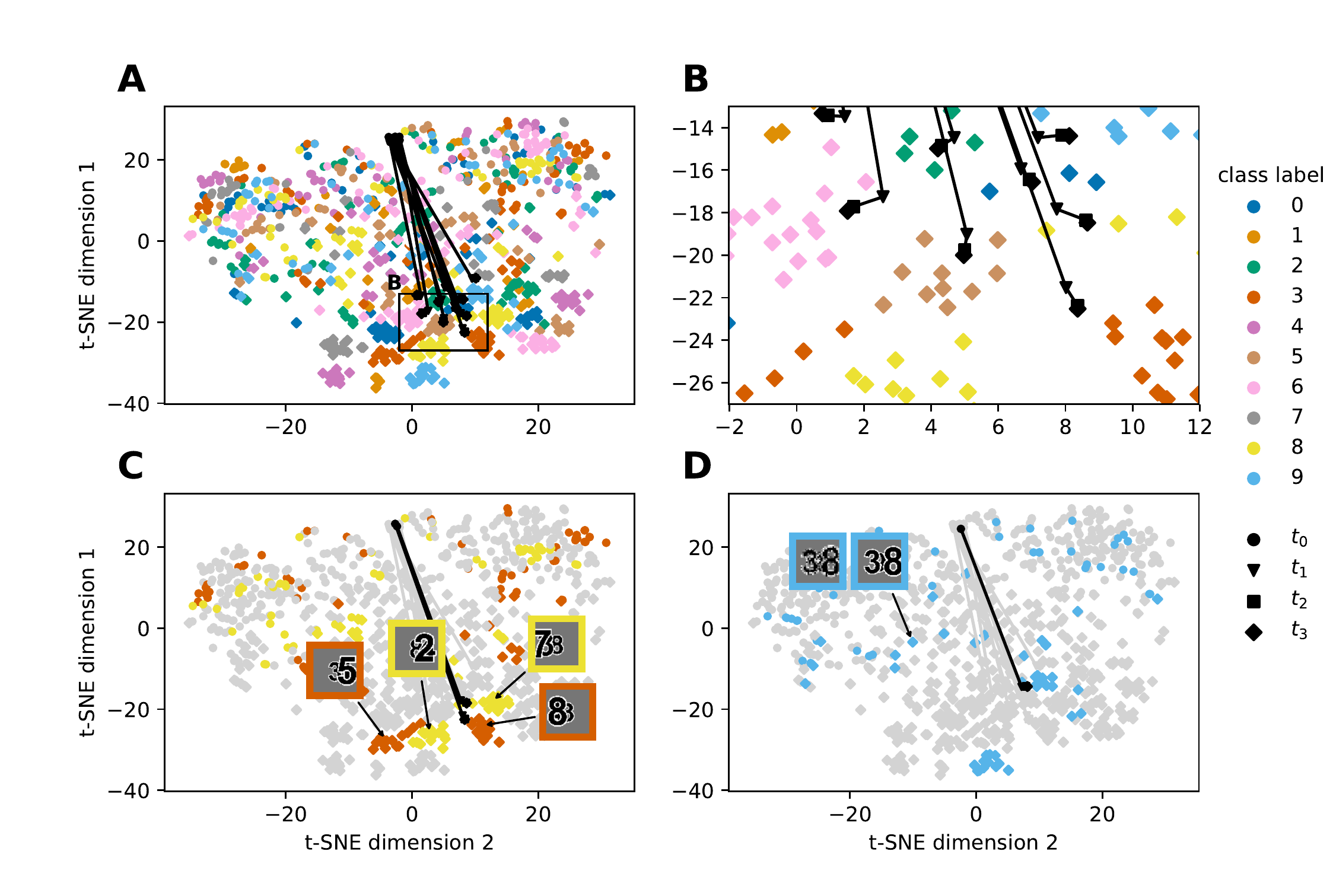}
\caption{t-SNE depiction of the network's representation of un-occluded stimuli (black) evolving in time. Time trajectories are shown as black lines, the colors represent clusters of different classes. See text for details.} 
\label{fig:tsne_depiction}
\end{figure}

To better understand how recurrent connections contribute to the performance gains, we consider the activation patterns $\mathbf{a}^{(t)}$ from the last hidden layer of the network for every time step $t$. 
We visualize the corresponding high-dimensional space using t-SNE \cite{maaten2008visualizing}, see Fig.~\ref{fig:tsne_depiction}. The black lines represent the activations caused by un-occluded stimuli evolving in time, henceforth called time-trajectories. The colored scatter-plots illustrate the activations corresponding to occluded stimuli of different classes at different time steps.

Interestingly, the t-SNE visualization reveals the internal representation at the first time step to be more intermingled where arguably very similar stimuli across classes are placed close to each other. At later time steps, however, the representation becomes well-separated and classes have seemingly no overlap.
As seen in Fig.~\ref{fig:tsne_depiction} A, the representation of the un-occluded stimuli moves towards a corresponding class-cluster at $t_3$. The first time step accounts for most of the distance travelled, followed by fine adjustments at $t_2$ and $t_3$, see detailed view in Fig.~\ref{fig:tsne_depiction} B. 
Class-clusters at $t_3$ are actually cluster pairs, corresponding to the occluders appearing mainly on the left or on the right (Fig.~\ref{fig:tsne_depiction} C). Indeed, when we tested the network with digits that were occluded from one side only, this structure disappeared and every digit was represented by just a single cluster at $t_3$. Activation patterns that do not fall close to a cluster tend to be dominated by the occluders. This is illustrated in Fig.~\ref{fig:tsne_depiction} D by a sample of class 9 (blue), shown in high resolution and as seen by the network.

\begin{figure}[hbtp]
\centering
\includegraphics[width=0.49\textwidth]{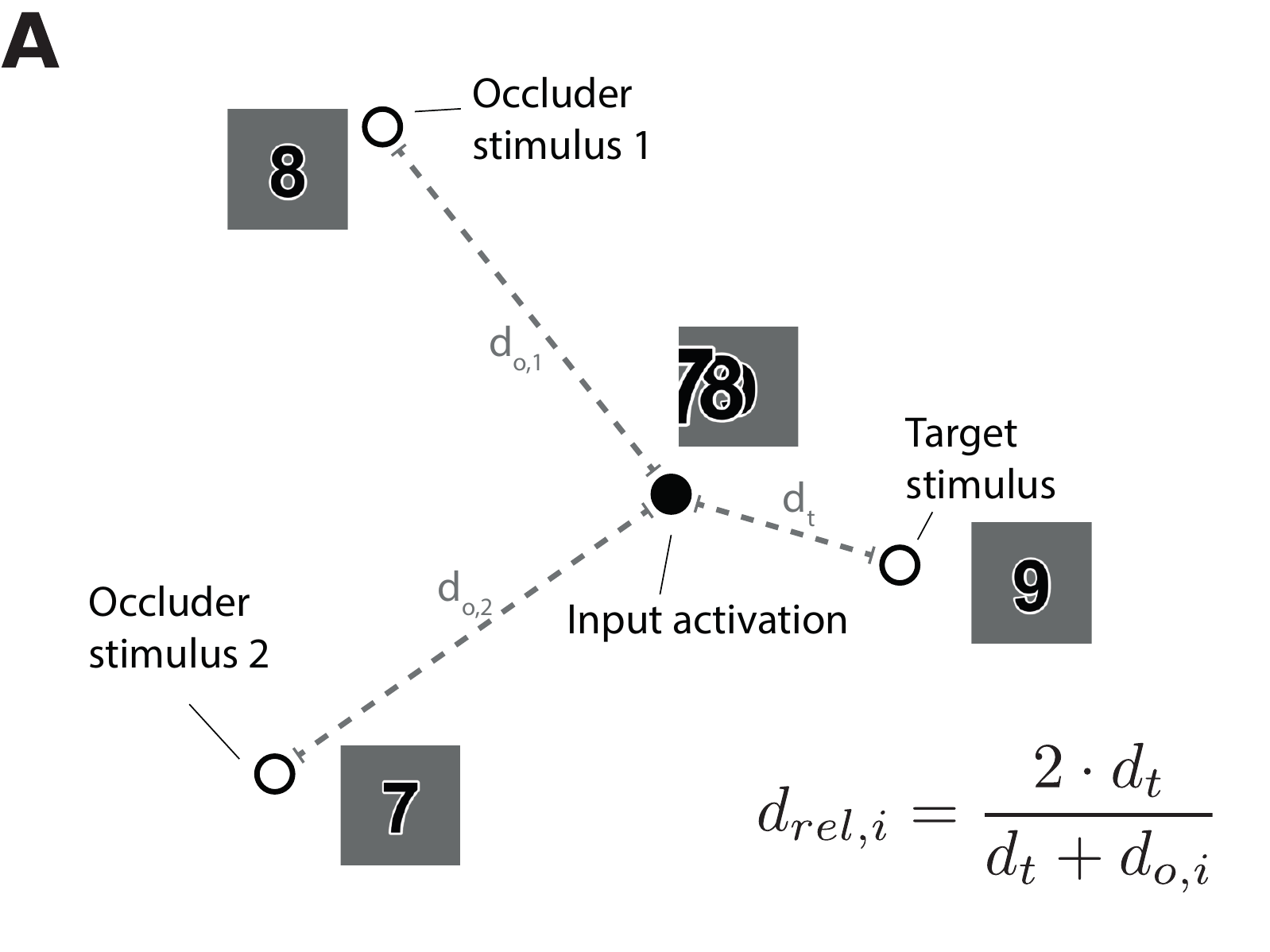}
\includegraphics[width=0.49\textwidth]{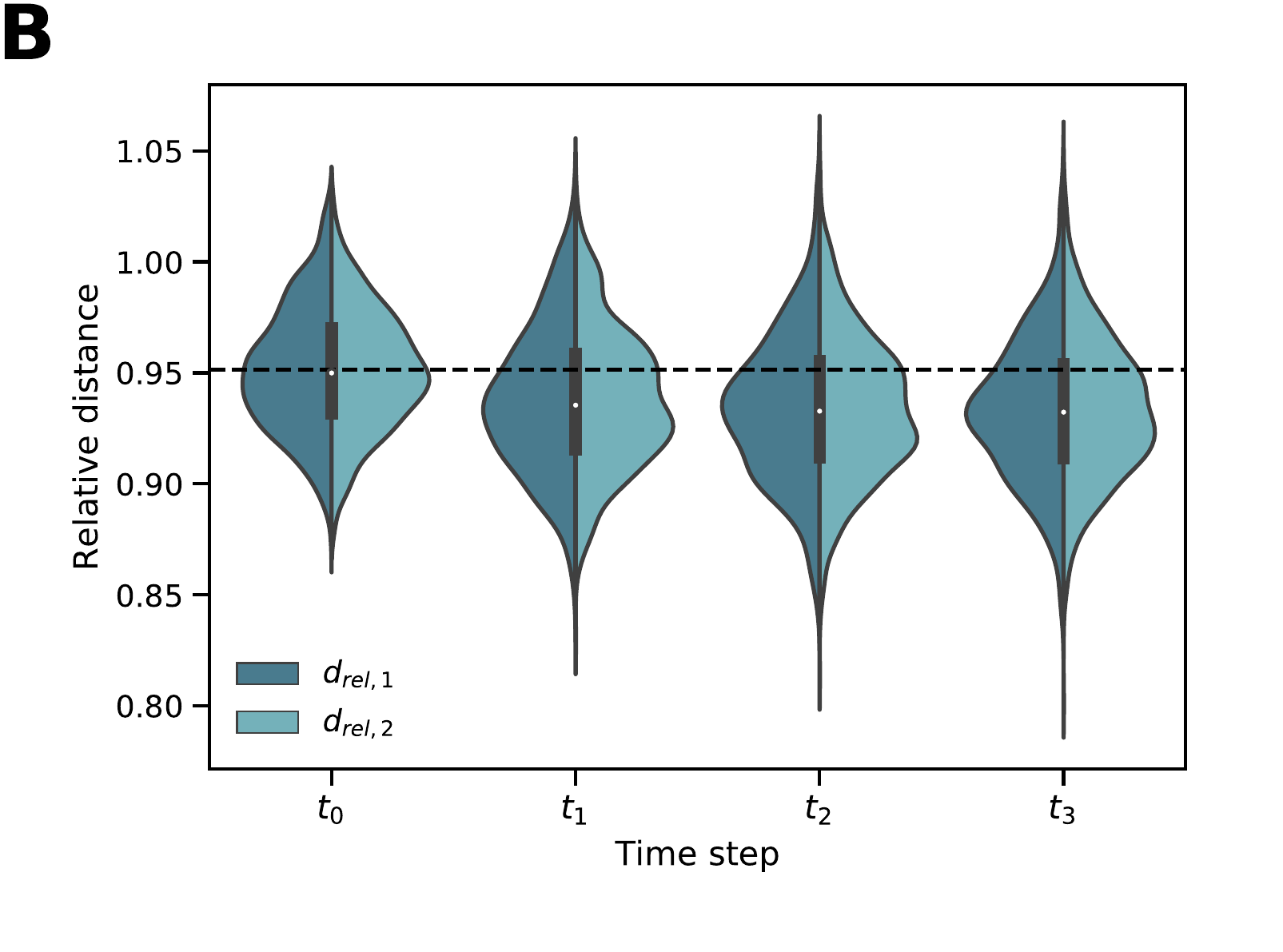}
\caption{Analysis of the internal representation of occluded stimuli shows discounting of occluders due to recurrent processing. (A) We define a relative distance measure to quantify if the activation of a stimulus is relatively closer to the un-occluded target compared to the occluder stimuli. Values below one indicate relative proximity to the target. (B) Violin plot displays the relative distances to occluder 1 and 2 at different time steps for stimuli occluded from the left. Dashed line represents mean of the distribution at $t_0$.}
\label{fig:cluster_analysis}
\end{figure}

We hypothesize that the recurrent connections help to discount the occluders by keeping the internal representation of the input close to that of a pure, un-occluded target stimulus. To test this, we compare the distances between activations caused by stereo-digit input, the un-occluded target and un-occluded occluders (Fig.~\ref{fig:cluster_analysis}). The resulting relative distances reveal that the representation of the input grows closer to the target stimulus relative to the occluder stimuli over time. This finding is consistent with the idea that the recurrent connections allow the network to discount the occluders and is also backed up by the finding that the sum of recurrent weights (lateral and top-down) becomes slightly negative during training (not shown).

\section{Discussion}
We studied if and how recurrent connections benefit occluded object recognition. Similar to \cite{spoerer2017recurrent}, but for a more realistic dataset we found that recurrent models significantly outperformed their non-recurrent counterparts for a near-equal amount of learnable parameters. Of the feedforward ensemble \emph{B-F} performed best on the given task suggesting that additional convolutional filters are more helpful than larger filters.
Contrary to the findings in \cite{spoerer2017recurrent} \emph{B-K} did not perform better than the standard \emph{B} model. The reported advantage of larger receptive fields might be linked to stimuli, where certain irregularities of the occluders only become obvious at larger scales.
Of the recurrent network ensemble \emph{BT} performed worst, suggesting that lateral connections are particularly important. As information has to pass through a convolution, maxpooling and back through a deconvolution instead of being directly transferred within the same layer, information may suffer in quality. In contrast, the \emph{BL} model takes advantage of the lateral shortcuts to pass information through time without utilizing more abstract features. The combination of both models (\emph{BLT}) performs best in all runs.

As previous work has shown, similar recurrent architectures also outperform parameter matched control models when no occlusion is present \cite{liang2015recurrent}. This is corroborated by studies that have investigated how object information is unfolding over time in the brain \cite{brincat2006dynamic}. Therefore, some level of recurrent connectivity in artificial neural networks might be beneficial even for standard object recognition tasks.

The significant performance gains for stereoscopic input can be explained by the fact that the additional input channel introduces a new perspective of the scene, thus giving the network more information about the occluded target. Qualitatively, the results of the statistical network comparisons resemble the ones obtained for monocular stimuli. Interestingly, however, the performance difference between recurrent and feedforward models was substantially higher for stereoscopic stimuli. 
The on average slightly negative weights of the recurrent connections might contribute to inhibiting or discounting occluders. With the networks dynamics being determined by the ReLU activation function a slight bias towards inhibitory weights might also be key to keep activations centered around the non-linearity.

For recurrent architectures, the probability distribution over possible outcomes has been shown to evolve with time. For challenging stimuli the recurrent dynamics are able to revise the best guess for the target.
In line with \cite{oreilly2013recurrent,wyatte2012limits}, we hypothesized that some of the missing information from occluded regions of the input image is recovered over time by the recurrent connections. The evolution of the output and our qualitative t-SNE analysis supports this hypothesis: At the first time step representations are more intermingled and spread out. Visually very similar stimuli across classes are represented close to each other. At later time steps the representation has evolved in such a way that the classes are well-separated.

The visualization revealed that the recurrent network distinguishes which side of the target stimulus is occluded. This is indicated by the representation of one digit being separated into two well-separated clusters. We conjecture that the network learns the left and right half of the un-occluded stimulus and therefore is able to ignore distractors on the opposing sides. The fragmentation nevertheless allows for high accuracy classification by the final readout layer, since the subsequent fully connected layer is able to classify correctly as long as the activation patterns remain linearly separable.

Investigating the internal representation we could also show that activations of occluded stimuli over time are grouped together with the activation caused by the un-occluded target stimulus relative to the un-occluded occluder digits. The representation being closer to the un-occluded target hints at the recurrent connections playing an important role in discounting the occluders.

In conclusion, recurrent convolutional neural networks have been shown to outperform feedforward networks at occluded object recognition. Building on previous work where parts of the target object were deleted \cite{oreilly2013recurrent} or occluded \cite{spoerer2017recurrent} we could show that the same performance advantages exist for a more realistic 3D occlusion scenario with stereoscopic input. In fact, the advantages were even greater for the more realistic stereoscopic input compared to monocular input.
Furthermore, our analysis revealed how recurrent connections revise the network's output over time, sometimes correcting an incorrect initial output after the first feedforward pass through the network. Future work should investigate whether these advantages generalize to larger and more complex network architectures than we considered in this work. Given the better performance and greater biological plausibility of recurrent network architectures, they deserve more detailed study.

\bibliographystyle{splncs04}
\bibliography{mybibliography}

\end{document}